\documentclass[11pt]{article}

\usepackage[preprint]{acl}

\usepackage{times}
\usepackage{latexsym}

\usepackage[T1]{fontenc}

\usepackage[utf8]{inputenc}

\usepackage{microtype}

\usepackage{inconsolata}

\usepackage{graphicx}

\usepackage{multirow} 
\usepackage[linesnumbered,ruled,vlined]{algorithm2e}

\usepackage{xspace}
\usepackage{subcaption}
\usepackage{bm}
\usepackage{pifont} 
\usepackage{amsmath} 
\usepackage{makecell}
\usepackage{enumitem}
\usepackage{booktabs}
\usepackage{amsthm}
\usepackage{amssymb}
\usepackage[table]{xcolor}
\usepackage{xcolor}

\newcommand{\ourmethod}{MCA$^2$\xspace}
\newtheorem{definition}{Definition}
\SetKwInOut{Param}{Parameters}

\title{Beyond a Single Perspective: Text Anomaly Detection \\with Multi-View Language Representations}

\author{
\textbf{Yixin Liu\textsuperscript{1}\thanks{These authors contributed equally.}}, 
\textbf{Kehan Yan\textsuperscript{2}\footnotemark[\value{footnote}]}, 
\textbf{Shiyuan Li\textsuperscript{1}\footnotemark[\value{footnote}]}, 
\textbf{Qingfeng Chen\textsuperscript{2}\thanks{Corresponding author.}}, 
\textbf{Shirui Pan\textsuperscript{1}} \\
 \textsuperscript{1}School of Information and Communication Technology, Griffith University,  Australia,
 \\
 \textsuperscript{2}School of Computer, Electronics and Information, Guangxi University, China
 \\
 \{{yixin.liu},  {s.pan}\}@griffith.edu.au, 2413301048@st.gxu.edu.cn \\
 qingfeng@gxu.edu.cn, 
 shiyuan.li@griffithuni.edu.au 
}

\begin{document}
\maketitle
\begin{abstract}
Text anomaly detection (TAD) plays a critical role in various language-driven real-world applications, including harmful content moderation, phishing detection, and spam review filtering. While two-step ``embedding–detector'' TAD methods have shown state-of-the-art performance, their effectiveness is often limited by the use of a single embedding model and the lack of adaptability across diverse datasets and anomaly types. To address these limitations, we propose to exploit the embeddings from multiple pretrained language models and integrate them into \ourmethod, a multi-view TAD framework. \ourmethod adopts a multi-view reconstruction model to effectively extract normal textual patterns from multiple embedding perspectives. To exploit inter-view complementarity, a contrastive collaboration module is designed to leverage and strengthen the interactions across different views. Moreover, an adaptive allocation module is developed to automatically assign the contribution weight of each view, thereby improving the adaptability to diverse datasets. Extensive experiments on 10 benchmark datasets verify the effectiveness of \ourmethod against strong baselines. The source code of \ourmethod is available at \url{https://github.com/yankehan/MCA2}. 
\end{abstract}

\section{Introduction}
Text anomaly detection (TAD) is a fundamental research problem that aims to identify anomalous or suspicious textual instances that deviate from normal patterns~\cite{pang2021deep,cao2025tad}. With the ever-increasing volume of digital text data, TAD plays a crucial role in various real-world applications. For instance, TAD helps detect abusive or threatening messages to maintain the safety and integrity of social platforms~\cite{fortuna2018survey}. Meanwhile, identifying anomalous product reviews via TAD is essential for ensuring the reliability of e-commerce ecosystems~\cite{chino2017voltime}. Due to its wide range of applications, in recent years, TAD has attracted increasing research attention in the research community of natural language processing~(NLP)~\cite{li2024nlp,cao2025tad,cao2025text,manolache2021date}. 

Recent advances in large language models (LLMs) have greatly improved their representation capabilities, enabling them to generate high-quality contextualized embeddings that capture rich semantic content and syntactic patterns in textual data~\cite{li2026ofa,cao2025tad,bai2025qwen2,neelakantan2022text}. By integrating high-quality textual embeddings with various anomaly detectors, \textbf{embedding-based methods} have demonstrated promising effectiveness in addressing the TAD problem~\cite{cao2025tad}. Typically, embedding-based methods follow a two-step pipeline: First, a text embedding model (e.g., BERT~\cite{devlin2019bert} or OpenAI’s embedding model~\cite{openai2024embedding}) encodes the text into numerical embeddings. Then, anomaly detection algorithms designed for vectorized data (e.g., LOF~\cite{breunig2000lof} and iForest~\cite{liu2008isolation}) are applied to detect anomalies based on the embeddings. Due to the powerful representation capability of LLM-generated embeddings, these embedding-based methods achieve state-of-the-art performance in TAD tasks and even outperform many end-to-end approaches~\cite{li2024nlp}.

\begin{figure}[t]
    \centering
    \begin{subfigure}[b]{1\linewidth}
        \centering
        \includegraphics[width=\linewidth]{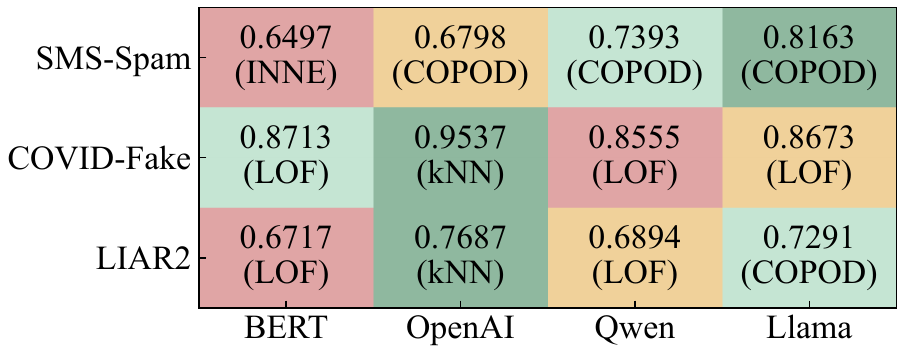}
        \caption{AUROC of different embedding models with (best detector); colors indicate {\color[RGB]{143, 184, 159}1st}, {\color[RGB]{197, 229, 211}2nd}, {\color[RGB]{240, 209, 154}3rd}, and {\color[RGB]{224, 165, 165}4th} ranks.}
        \label{fig:intro_heatmap}
    \end{subfigure}
    \begin{subfigure}[b]{1\linewidth}
        \centering
        \includegraphics[width=\linewidth]{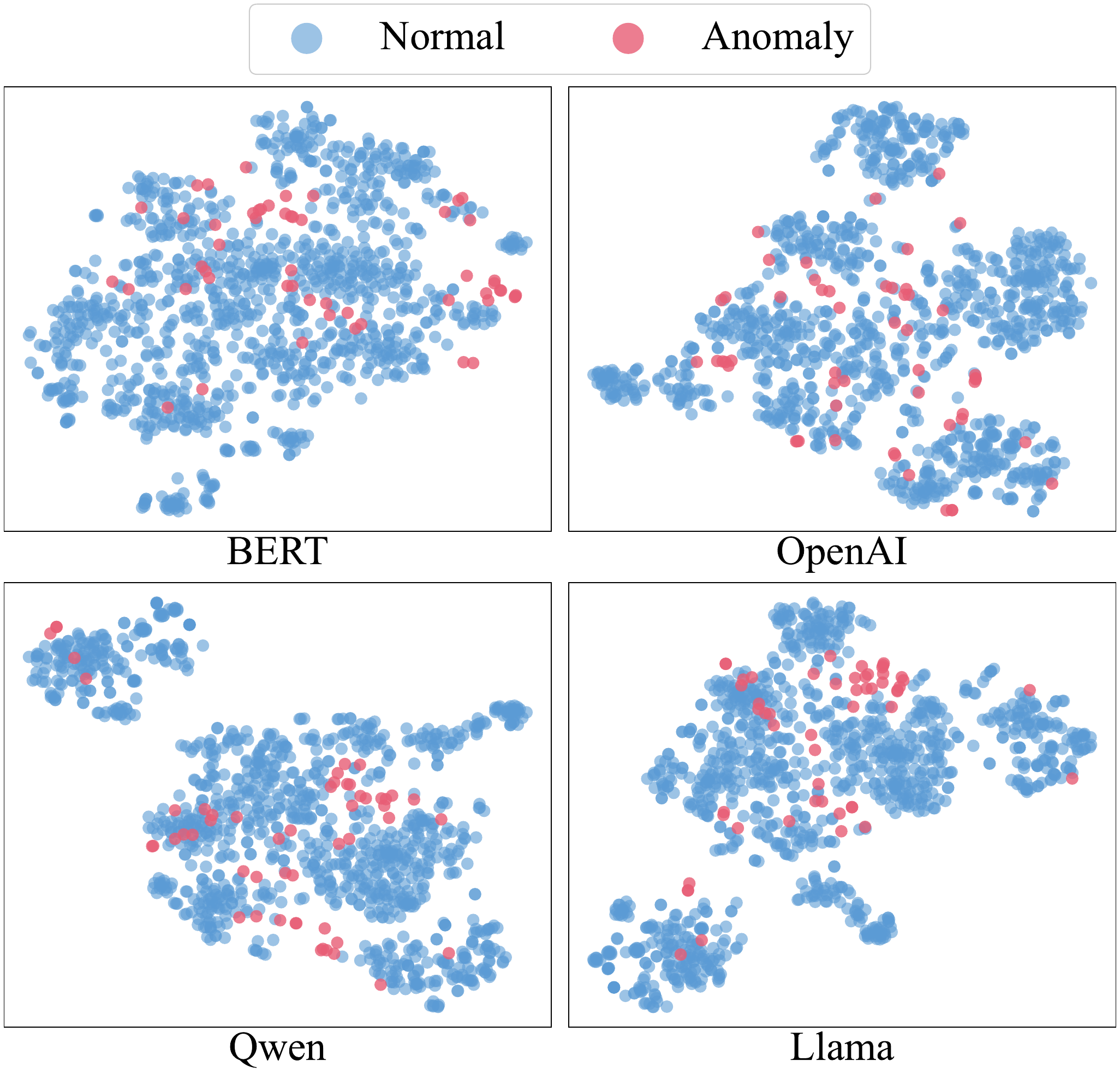}
        \caption{Visualization of embedding distributions via t-SNE.}
        \label{fig:intro_tsne}
    \end{subfigure}
    \caption{(a) Performance comparison of different embedding models with the best detectors. (b) Visualization of embeddings on COVID-Fake dataset.}
    \label{fig:intro}
\end{figure}

Despite their remarkable performance, these embedding-based methods are usually built upon embeddings produced by \textit{a single text embedding model} and rely on \textit{a particular detector} to conduct anomaly detection, which leads to several inherent limitations. Firstly, a single embedding model is inevitably biased toward the distribution and linguistic characteristics of its pretraining corpus, which prevents it from providing universally robust representations in TAD scenarios with diverse domains, anomaly types, and textual styles. As a result, different embedding models exhibit varying performance across datasets, and no single embedding model consistently emerges as the overall winner, as demonstrated in Figure~\ref{fig:intro_heatmap} (source data from recent benchmark~\cite{cao2025tad}). Moreover, due to the diverse embedding distributions produced by different embedding models (Figure~\ref{fig:intro_tsne} gives an example), it is non-trivial to determine in advance which detector will be most compatible with the corresponding distribution for a given dataset. Consequently, we have to evaluate all possible embedding–detector combinations to determine the best-performing one, which may be costly and impractical in real-world cases. Motivated by these issues, a natural question arises: \textit{\textbf{Can we develop a unified TAD framework that leverages embeddings from multiple models and integrates them to achieve better anomaly detection?}}

To answer this question, two pressing challenges need to be tackled. \textit{\textbf{Challenge~1:} How to coordinate multiple embeddings and fully exploit their complementary strengths?} As shown in Figure~\ref{fig:intro_tsne}, embeddings generated by different models capture the data from different perspectives and encode complementary information. Then, how to make them collaborate in a way that they can both mutually enhance and constrain each other is a critical challenge. \textit{\textbf{Challenge~2:} How to adaptively balance the contributions of different embeddings to achieve effective anomaly detection?} Due to the diversity of textual data and embedding models, the usefulness of different embeddings for anomaly detection often differs across datasets, which is proven in Figure~\ref{fig:intro_heatmap}. In this case, a unified TAD framework should also adaptively determine how much each embedding contributes based on the characteristics of the data. 

Motivated by these challenges, in this paper, we propose a novel \textbf{M}ulti-view TAD framework with \textbf{C}ontrastive \textbf{C}ollaboration and \textbf{A}daptive \textbf{A}llocation (\ourmethod for short). \ourmethod is built upon a multi-view reconstruction model for TAD, where embeddings from different models form multiple complementary views, equipped with two well-crafted modules to further exploit such multi-view information. More specifically, to tackle \textit{\textbf{Challenge~1}}, we design a \textit{contrastive collaboration module} that enforces distributional consistency among different views in the latent space. By maximizing the mutual information between different views, the latent distributions become better aligned and more structurally consistent, leading to more discriminative anomaly characteristics. In addition, the mutual information can be further exploited as an indicator of abnormality. To address \textit{\textbf{Challenge~2}}, we develop an \textit{adaptive allocation module} that automatically assigns appropriate importance to different views based on the characteristics of the data. This module not only allows \ourmethod to adapt across various datasets, but also provides sample-level adaptiveness for more precise anomaly detection. To sum up, the contributions of this paper are threefold:

\begin{itemize}[leftmargin=*]
    \item \textbf{New Paradigm:} Going beyond single embedding model-based methods, we take the first step to leverage multiple embedding models to capture complementary information for TAD.
    \item \textbf{Novel Method:} We propose \ourmethod, a multi-view TAD framework that adaptively allocates the contributions of different embeddings and enforces contrastive collaboration across multiple views.
    \item \textbf{Extensive Experiments:} Extensive experiments on ten real-world benchmark datasets demonstrate that our method achieves superior anomaly detection performance compared with state-of-the-art approaches.
\end{itemize}

\section{Related Work}
In this section, we provide a brief summary of the studies in three key areas: text embedding, text anomaly detection, and multi-view anomaly detection. Detailed reviews are provided in Appendix~\ref{app:rw}. 

\noindent\textbf{Text Embedding} techniques aim to map textual data into vectorized representations (a.k.a. embeddings) that capture semantic and syntactic information. Early methods involve TF-IDF~\cite{salton1988term} or Word2Vec~\cite{mikolov2013efficient} to learn text representations. Then, BERT~\cite{devlin2019bert} and its advanced models show strong representation capability through large-scale pretraining. In the era of LLMs, text embeddings generated by billion-parameter models (e.g., OpenAI models~\cite{openai2024embedding} and Qwen~\cite{zhang2025qwen3}) have become more expressive and powerful, providing high-quality embeddings for various downstream tasks, including anomaly detection. 

\noindent\textbf{Text Anomaly Detection (TAD)} aims to identify textual instances that deviate from dominant normal data~\cite{li2024nlp}. Existing TAD methods can be divided into two categories. End-to-end methods perform anomaly detection in a unified manner by directly predicting abnormality from raw textual inputs~\cite{manevitz2007one,ruff2019self,manolache2021date,das2023few}. Embedding-based methods first convert text into dense embeddings using text embedding models, and then apply anomaly detectors (such as LOF~\cite{breunig2000lof}, KNN~\cite{ramaswamy2000efficient}, COPOD~\cite{li2020copod}, iForest~\cite{liu2008isolation}, and LUNAR~\cite{goodge2022lunar}) for detection. Although empirical evidence~\cite{li2024nlp,cao2025tad} shows that embedding-based methods often achieve state-of-the-art performance, they typically rely on a single embedding model, which makes them less robust when facing diverse datasets and anomaly types.

\noindent \textbf{Multi-View Anomaly Detection} focuses on identifying anomalous samples in multi-view data, e.g., image data represented by multiple views like color and shape feature descriptors. Early studies detect anomalies based on multi-view clustering~\cite{marcos2013clustering,liu2012using}. Recent studies, such as NCMOD~\cite{cheng2021neighborhood} and RCPMOD~\cite{wang2024regularized}, employ unsupervised deep learning models to detect anomalies. 
Despite their success in multi-view visual data, how to conduct multi-view anomaly detection for high-dimensional textual data remains open.

\section{Problem Definition}
\begin{figure*}
\centering
\includegraphics[width=1\linewidth]{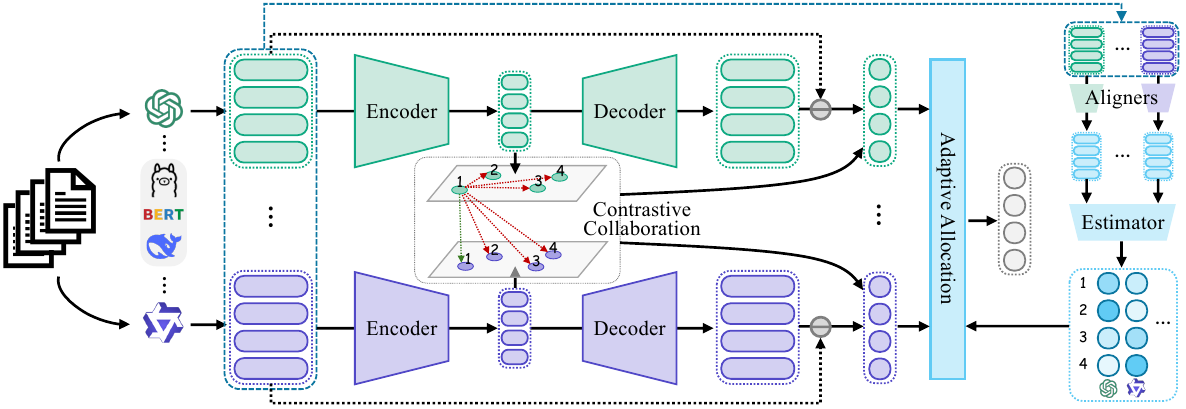}
\caption{
Overall framework of \ourmethod. We illustrate the case of two views ({\color[RGB]{77,166,132} OpenAI} and {\color[RGB]{78,70,187} Qwen}) as an example.}
\label{fig:framework}
\end{figure*}

To leverage the embeddings generated by multiple embedding models for TAD, we formulate TAD as a multi-view anomaly detection problem. 

\begin{definition}[Multi-view Text Anomaly Detection]
Let $\mathcal{D}=\{x_i\}_{i=1}^{N}$ be a text dataset consisting of $N$ documents. Each instance $x_i$ is associated with a label $y_i \in \{0,1\}$,
where $y_i = 0$ denotes a normal sample and $y_i = 1$ denotes an anomalous sample. Given $K$ large language embedding models $\{f_k(\cdot)\}_{k=1}^{K}$, each text sample $x_i$ is mapped into $K$ embedding views:

\begin{equation}
\mathbf{v}_i^{(k)} = f_k(x_i), \quad k = 1,2,\dots,K.
\label{eq:mv_embed}
\end{equation}

Thus, each instance is represented as a multi-view embedding set
$\mathcal{V}_i = \{\mathbf{v}_i^{(1)},\dots,\mathbf{v}_i^{(K)}\}$. 
The goal of multi-view text anomaly detection is to learn an
anomaly scoring function
\begin{equation}
s(\mathcal{V}_i): \mathcal{V}_i \rightarrow \mathbb{R},
\label{eq:score_func}
\end{equation}
such that anomalous instances receive higher scores than normal
ones. A final decision is obtained by thresholding $s(\mathcal{V}_i)$.
\end{definition}

Considering that anomalies are difficult to obtain in real-world scenarios, we follow a one-class anomaly detection setting in this paper, consistent with existing benchmarks~\cite{li2024nlp}: The training dataset is defined as $\mathcal{D}_{\text{train}} = \{x_i \mid y_i = 0\}$, where only normal samples are available. The test set is $\mathcal{D}_{\text{test}} = \{x_j \mid y_j \in \{0,1\}\}$, which contains both normal and anomalous samples.

\section{Methodology}
In this section, we introduce the proposed multi-view TAD framework, \ourmethod, in detail. As illustrated in Figure~\ref{fig:framework}, \ourmethod employs an autoencoder-based \textit{multi-view reconstruction model} as the backbone of TAD model (Section~\ref{subsec:recon}). To exploit the complementary information across different views, we introduce an \textit{inter-view contrastive collaboration module} that maximizes the consistency among views (Section~\ref{subsec:cc}). Meanwhile, we introduce an \textit{adaptive view contribution allocation module} to dynamically assign the contribution of different views in indicating abnormality (Section~\ref{subsec:aa}). %

\subsection{Multi-view Reconstruction TAD Model}\label{subsec:recon}

To achieve effective TAD with embeddings from diverse models, the core is to find a universal anomaly detection paradigm that captures the data distributions of diverse and heterogeneous embeddings. Under the one-class setting, the lack of anomalous training samples further highlights the need for an unsupervised anomaly detection paradigm. Motivated by the \textit{reconstruction assumption}, i.e., a reconstruction model trained on normal data can well reconstruct normal samples while yielding large reconstruction errors for anomalous ones, a promising solution is to conduct anomaly detection under the reconstruction paradigm. As autoencoder-based anomaly detection models have proven to be effective in various data modalities (e.g., images~\cite{zhou2017anomaly}, time series~\cite{zamanzadeh2024deep}, and graphs~\cite{ding2019deep}) by modeling diverse data distributions, we build our multi-view TAD framework on a reconstruction-based backbone.

Considering the differences between embeddings generated by different language models (i.e., data views), we build an independent autoencoder for each view to perform reconstruction. Formally, for the data of the $k$-th view $\mathcal{V}^{(k)}=\{\mathbf{v}_1^{(k)}, \cdots, \mathbf{v}_N^{(k)}\}$, an MLP-based autoencoder is employed to model this view:
\begin{equation}
\mathbf{z}_i^{(k)} = \mathrm{Enc}^{(k)}(\mathbf{v}_i^{(k)}), \quad
\widehat{\mathbf{v}}_i^{(k)} = \mathrm{Dec}^{(k)}(\mathbf{z}_i^{(k)}),
\label{eq:ae}
\end{equation}

\noindent where $\mathrm{Enc}^{(k)}(\cdot)$ and $\mathrm{Dec}^{(k)}(\cdot)$ denote 
the encoder and decoder of the $k$-th view, respectively, $\mathbf{z}_i^{(k)}$ is the latent representation, and $\widehat{\mathbf{v}}_i^{(k)}$ is the reconstructed embedding of $\mathbf{v}_i^{(k)}$. The autoencoder can be optimized by minimizing the reconstruction loss between $\widehat{\mathbf{v}}_i^{(k)}$ and $\mathbf{v}_i^{(k)}$:
\begin{equation}
\mathcal{L}_{i,rec}^{(k)} 
=  
\left\|\mathbf{v}_i^{(k)} - \widehat{\mathbf{v}}_i^{(k)}\right\|_2^2.
\label{eq:l_rec}
\end{equation}

Once the model is well trained, the reconstruction error of each view can be used to measure the abnormality of the corresponding samples:

\begin{equation}
s_{i,rec}^{(k)} = 
\left\|\mathbf{v}_i^{(k)} - \widehat{\mathbf{v}}_i^{(k)}\right\|_2^2,
\label{eq:s_rec}
\end{equation}

\noindent where $s_{i,rec}^{(k)}$ denotes the reconstruction-based anomaly score of the $i$-th sample in the $k$-th view, computed as the squared $\ell_2$ reconstruction error. Owing to the strong capability of autoencoders in modeling various normal data distributions, the learned anomaly score can effectively indicate the abnormality of each sample from an intra-view perspective, reflecting the view-specific characteristics of the corresponding embedding space. 

\subsection{Inter-View Contrastive Collaboration}\label{subsec:cc}

Although the reconstruction-based backbone can capture intra-view information for anomaly detection, it may overlook the crucial inter-view dependencies and consistency. Since different embedding models can capture different aspects of textual semantics, the data from different views can be complementary. To further leverage such mutual complementarity to enhance multi-view TAD, we propose an inter-view contrastive collaboration mechanism that encourages different views to collaborate and complement each other. Our core idea is to maximize the mutual information between the representations of the same sample across different views, thereby aligning their distributions and improving the quality of the latent representations. More importantly, as the inter-view matching patterns can also expose abnormal behaviors, the mutual information can serve as an indicator of abnormality. This collaboration-based abnormality measurement provides a supplement to the reconstruction-based intra-view anomaly scores.

Since each pair of views can provide complementary information to one another, contrastive collaboration is conducted over all possible view pairs. Formally, given the latent representation sets of the $j$-th and $k$-th views, i.e., $\mathcal{Z}^{(j)}=\{\mathbf{z}_i^{(j)}\}_{i=1}^N$ 
and $\mathcal{Z}^{(k)}=\{\mathbf{z}_i^{(k)}\}_{i=1}^N$, we adopt an InfoNCE contrastive loss to enhance inter-view collaboration:

\begin{equation}
\mathcal{L}_{i,con}^{(j,k)}
= -\log p_i^{(j,k)},
\label{eq:l_con}
\end{equation}
\begin{equation}
p_i^{(j,k)}
=
\frac{
e^{s(\mathbf{z}_i^{(j)},\mathbf{z}_i^{(k)})/\tau}
}{
\sum_{\substack{m\neq i}} e^{s(\mathbf{z}_i^{(j)},\mathbf{z}_m^{(k)})/\tau}
+
\sum_{\substack{n\neq i}} e^{s(\mathbf{z}_i^{(j)},\mathbf{z}_n^{(j)})/\tau}
},
\label{eq:p_match}
\end{equation}

\noindent where $p_i^{(j,k)}$ denotes the matching probability of the $i$-th 
sample between the $j$-th and $k$-th views, $s(\cdot,\cdot)$ is the cosine similarity, and $\tau$ is a temperature hyperparameter. Note that we incorporate both cross-view and intra-view samples as negative instances, which helps learn more discriminative latent representations. In practice, we conduct contrastive learning in a mini-batch manner to ensure efficient optimization and stable training on large-scale datasets.

Trained with the contrastive loss, the model can learn the matching patterns across different views. That is to say, normal samples tend to exhibit strong and consistent cross-view correspondence due to the constraint imposed by the contrastive collaboration mechanism. On the other hand, an anomalous sample may break this cross-view consistency, making the matching probability a meaningful indicator of abnormality. Based on this property, we can obtain the contrastive anomaly score $s_{i,con}^{(k)}$ of the $i$-th sample in the $k$-th view by aggregating its matching probabilities with all the other views:

\begin{equation}
s_{i,con}^{(k)}
=
-\frac{1}{K-1}
\sum_{\substack{j=1 , \ j\neq k}}^{K}
\log p_i^{(j,k)}.
\label{eq:s_con}
\end{equation}

While $s_{i,rec}^{(k)}$ measures the abnormality from an intra-view perspective, $s_{i,con}^{(k)}$ provides a complementary measurement from an inter-view perspective, which improves the use of multi-view information for more accurate anomaly identification.

\subsection{Adaptive View Contribution Allocation}\label{subsec:aa}

After the reconstruction and contrastive collaboration modules produce the anomaly score of each sample at each view, the remaining problem is how to fuse them into a unified anomaly score. A naive solution is to simply aggregate the anomaly scores from different views with equal weights; however, due to the heterogeneous representational capabilities of embedding models and their varying adaptability to a specific dataset, treating all views equally may lead to suboptimal fusion, where informative views are diluted and less reliable ones are overweighted. In this case, a more desirable strategy is to adaptively weight different views in a data-driven manner. 
To achieve this goal, we incorporate an \textit{adaptive view contribution allocation} module into \ourmethod to automatically determine the importance of each view. This module takes the multi-view embeddings as input and outputs allocation weights for different views, which are used to guide the fusion of anomaly scores. 

\noindent \textbf{Weight Estimation.} Considering that the dimensions and semantics of different views are heterogeneous, in the first step, we perform an alignment to map them into a shared space. Due to its effectiveness in reducing dimensionality and retaining the principal structural information, we employ PCA algorithm~\cite{mackiewicz1993principal} as the aligner for each view. Concretely, for each view of data $\mathcal{V}^{(k)}=\{\mathbf{v}_1^{(k)}, \cdots, \mathbf{v}_N^{(k)}\}$, we stack them into a matrix $\mathbf{V}^{(k)} \in \mathbb{R}^{N\times d_k}$ and apply a PCA transformation, i.e., $\widetilde{\mathbf{V}}^{(k)} = \mathrm{PCA}(\mathbf{V}^{(k)})$, where $\widetilde{\mathbf{V}}^{(k)} \in \mathbb{R}^{N\times d}$ denotes the aligned feature of the $k$-th view, and all views are projected into the same $d$-dimensional space to ensure dimensional consistency. Since PCA sorts the components by their importance (i.e., explained variance), the projected representations become unified at the variance-structure level across different views. 

After that, we estimate the contribution weights for different views with a lightweight neural network. Given a sample $x_i$, we extract all its aligned feature vectors $\{\widetilde{\mathbf{v}}^{(k)}_i\}_{k=1}^{K}$ by taking the $i$-th row from each $\widetilde{\mathbf{V}}^{(k)}$, and then an MLP-based estimator is applied to generate contribution scores:

\begin{equation}
{w'}_i^{(k)} = \mathrm{MLP}\!\left(\widetilde{\mathbf{v}}_i^{(k)}\right),
\quad
w_i^{(k)} = 
\frac{\sigma\!\left({w'}_i^{(k)}\right)}
{\sum\limits_{j=1}^{K}\sigma\!\left({w'}_i^{(j)}\right)},
\label{eq:weight}
\end{equation}

\noindent where ${w'}_i^{(k)}$ denotes the estimated contribution score, $\sigma(\cdot)$ is the sigmoid function, and $w_i^{(k)}$ is the normalized allocation weight of the $k$-th view 
for sample $x_i$. 

\noindent \textbf{Anomaly Scoring.} With the allocation weights, we can now calculate the anomaly score of $x_i$ by aggregating the view-specific scores from the reconstruction and contrastive modules. Concretely, the final anomaly score $s(x_i)$ is computed as:

\begin{equation}
s(x_i) = 
\sum_{k=1}^{K} 
w_i^{(k)}
\big(
\alpha\, s_{i,\text{rec}}^{(k)} 
+ 
\beta\, s_{i,\text{con}}^{(k)}
\big),
\label{eq:final_score}
\end{equation}

\noindent where $\alpha$ and $\beta$ are balance hyperparameters for reconstruction-based and 
contrastive-based scores, respectively. Note that the adaptive allocation is conducted at a fine-grained sample level rather than at a coarse dataset level, which enables the model to tailor the view contributions to each instance and leads to reliable anomaly identification.

\noindent \textbf{Two-Stage Training.} All the parameters in \ourmethod, including the detection model and the allocation module, are optimized by 
minimizing the overall loss function:

\begin{equation}
\mathcal{L}
=
\frac{1}{N}
\sum_{i=1}^{N}
\Big(
\sum_{k}
w_i^{(k)}\mathcal{L}_{i,\text{rec}}^{(k)}
+
\lambda
\sum_{j<k}
\omega_i^{(j,k)}
\mathcal{L}_{i,\text{con}}^{(j,k)}
\Big),
\label{eq:loss}
\end{equation}

\noindent where $\omega_i^{(j,k)}=({w_i^{(j)}+w_i^{(k)}})/{2}$ and $\lambda$ is a balance hyperparameter for two losses. 

While jointly optimizing all the parameters may be straightforward, it may lead to unstable training due to the coupling between the detection backbone and the allocation module. To ensure stable optimization of the entire framework, we adopt a {decoupled two-stage training strategy}. In the \textbf{first stage}, we freeze the parameters of the allocation module and enforce it to output uniform weights (i.e., $w = {1}/{K}$). This allows the encoders and decoders to fully learn robust feature reconstructions and cross-view collaboration. In the \textbf{second stage}, we freeze the detection model and only train the allocation module to assign appropriate view contributions for different samples based on reliable view-specific anomaly scores. 
The overall running algorithm of \ourmethod is given in Appendix~\ref{app:algo}, with complexity analysis given in Appendix~\ref{app:complex}.

\section{Experiments}
\subsection{Experiment Setup}
\begin{table*}[t]
\centering
\resizebox{1\linewidth}{!}{%
\begin{tabular}{l c c c c c c c c c c}
\toprule
\textbf{Methods} & \textbf{\shortstack{NLPAD-\\AGNews}} & \textbf{\shortstack{NLPAD-\\BBCNews}} & \textbf{\shortstack{NLPAD-\\MovieReview}} & \textbf{\shortstack{NLPAD-\\N24News}} & \textbf{\shortstack{TAD-\\EmailSpam}} & \textbf{\shortstack{TAD-\\SMSSpam}} & \textbf{\shortstack{TAD-\\OLID}} & \textbf{\shortstack{TAD-\\HateSpeech}} & \textbf{\shortstack{TAD-\\CovidFake}} & \textbf{\shortstack{TAD-\\Liar2}} \\
\midrule
CVDD & 0.6461 & 0.6150 & 0.4860 & 0.6443 & 0.8480 & 0.4499 & 0.5504 & 0.5108 & 0.7774 & 0.6646 \\
DATE & 0.7843 & 0.8026 & 0.4871 & 0.6609 & 0.9638 & \cellcolor{gray!20}\textbf{0.9670} & 0.5194 & 0.6009 & 0.7791 & 0.6900 \\
FATE & 0.8837 & 0.8221 & 0.5770 & 0.8770 & 0.7785 & 0.9518 & 0.5555 & 0.6774 & 0.8331 & 0.6424 \\
\midrule
BERT+LOF & 0.7432 & 0.9320 & 0.4959 & 0.6703 & 0.7530 & 0.6842 & 0.5123 & 0.4706 & 0.8524 & 0.6670 \\
BERT+DeepSVDD & 0.5558 & 0.5852 & 0.4507 & 0.4484 & 0.6200 & 0.5870 & 0.5220 & 0.4991 & 0.7147 & 0.5842 \\
BERT+ECOD & 0.6318 & 0.6912 & 0.4282 & 0.4969 & 0.6978 & 0.5675 & 0.5054 & 0.4899 & 0.7642 & 0.6091 \\
BERT+iForest & 0.6287 & 0.6844 & 0.4242 & 0.4808 & 0.6721 & 0.5715 & 0.4989 & 0.4920 & 0.7675 & 0.5904 \\
BERT+SO-GAAL & 0.4488 & 0.3100 & 0.4663 & 0.4140 & 0.4863 & 0.4111 & 0.4744 & 0.5232 & 0.6739 & 0.5154 \\
BERT+AE & 0.7197 & 0.8854 & 0.4650 & 0.5741 & 0.7585 & 0.6997 & 0.5120 & 0.4803 & 0.8275 & 0.6393 \\
BERT+VAE & 0.6778 & 0.7450 & 0.4387 & 0.5066 & 0.7228 & 0.6181 & 0.5092 & 0.4893 & 0.7685 & 0.6336 \\
BERT+LUNAR & 0.7654 & 0.9381 & 0.4647 & 0.6275 & 0.8443 & 0.7179 & 0.5305 & 0.5125 & 0.8492 & 0.6583 \\
OAI-L+LOF & 0.7879 & 0.9558 & 0.7292 & 0.7495 & 0.8339 & 0.7430 & 0.5614 & 0.5921 & 0.8523 & 0.7577 \\
OAI-L+DeepSVDD & 0.5019 & 0.5690 & 0.5132 & 0.6056 & 0.5831 & 0.4516 & 0.5299 & 0.4989 & 0.5727 & 0.4906 \\
OAI-L+ECOD & 0.6673 & 0.7225 & 0.4895 & 0.6216 & 0.9220 & 0.4238 & 0.5284 & 0.3465 & 0.8848 & 0.6280 \\
OAI-L+iForest & 0.5750 & 0.6211 & 0.5221 & 0.5746 & 0.8844 & 0.4933 & 0.5440 & 0.4612 & 0.7816 & 0.5654 \\
OAI-L+SO-GAAL & 0.3685 & 0.2359 & 0.4210 & 0.2920 & 0.1889 & 0.4643 & 0.4903 & 0.3399 & 0.3055 & 0.4643 \\
OAI-L+AE & 0.8916 & 0.9527 & 0.5942 & 0.7987 & 0.9381 & 0.5040 & 0.5504 & 0.6111 & 0.9520 & 0.7185 \\
OAI-L+VAE & 0.8514 & 0.7541 & 0.5248 & 0.7181 & 0.9257 & 0.4387 & 0.5302 & 0.3462 & 0.8945 & 0.6210 \\
OAI-L+LUNAR & 0.8998 & 0.9771 & 0.8258 & 0.8577 & 0.9828 & 0.7184 & 0.5730 & 0.7192 & 0.9651 & 0.7704 \\
\midrule
NCMOD (\textbf{OpenAIs}) & 0.7304 & 0.8451 & 0.6294 & 0.6861 & 0.9469 & 0.8199 & 0.6262 & 0.4929 & 0.8999 & 0.7169 \\
NCMOD (\textbf{Mixed}) & 0.7222 & 0.7642 & 0.5560 & 0.6508 & 0.9113 & 0.5207 & 0.5138 & 0.5047 & 0.8693 & 0.6807 \\
\midrule
RCPMOD (\textbf{OpenAIs}) & 0.7864 & 0.9778 & 0.7963 & 0.9570 & 0.8753 & 0.8169 & 0.5882 & 0.7132 & 0.8997 & 0.7213 \\
RCPMOD (\textbf{Mixed}) & 0.8249 & 0.9794 & 0.7249 & 0.9449 & 0.8204 & 0.7601 & 0.5305 & 0.5662 & 0.9021 & 0.6542 \\
\midrule
\ourmethod (\textbf{OpenAIs}) & \cellcolor{gray!20}\textbf{0.9484} & \cellcolor{gray!20}\textbf{0.9860} & \cellcolor{gray!20}\textbf{0.8381} & \cellcolor{gray!20}\textbf{0.9656} & \cellcolor{gray!20}\textbf{0.9895} & 0.8865 & \cellcolor{gray!20}\textbf{0.6355} & \cellcolor{gray!20}\textbf{0.7379} & 0.9531 & \cellcolor{gray!20}\textbf{0.7965} \\
\ourmethod (\textbf{Mixed}) & 0.9482 & 0.9752 & 0.7914 & 0.9575 & 0.9734 & 0.8143 & 0.4892 & 0.5647 & \cellcolor{gray!20}\textbf{0.9776} & 0.7287 \\
\bottomrule
\end{tabular}}
\caption{Main results on AUROC. Best results are highlighted in bold and shaded.}
\label{tab:main_res}
\end{table*}

\noindent\textbf{Datasets.} We conduct experiments on 10 public datasets from NLP-ADBench~\cite{li2024nlp} and TAD-Bench~\cite{cao2025tad}. Following the standard protocol in NLP-ADBench, we allocate 70\% of the normal instances for training. The remaining 30\% normal instances, together with all anomalous instances, form the test set. Details of the datasets are provided in Appendix~\ref{app:dset}.%

\noindent\textbf{Baselines.} We compare \ourmethod with three types of methods. \ding{182}~\textbf{End-to-end methods} include CVDD~\cite{ruff2019self}, DATE~\cite{manolache2021date}, and FATE~\cite{das2023few}. \ding{183}~\textbf{Embedding-based methods} first extract embeddings from a pretrained model (BERT~\cite{devlin2019bert} or OpenAI-large~\cite{openai2024embedding}) and apply an anomaly detector, selected from  LOF~\cite{breunig2000lof}, DeepSVDD~\cite{ruff2018deep}, ECOD~\cite{li2022ecod}, iForest~\cite{liu2008isolation}, SO-GAAL~\cite{liu2019generative}, AE~\cite{aggarwal2016introduction}, VAE~\cite{kingma2013auto,burgess2018understanding}, and LUNAR~\cite{goodge2022lunar}. \ding{184}~\textbf{Multi-view methods}, including NCMOD~\cite{cheng2021neighborhood} and RCPMOD~\cite{wang2024regularized}, are implied on the same embedding sets as \ourmethod. %

\noindent\textbf{Evaluation and Implementation.} We report AUROC and AUPRC as the main metrics. For all methods, we report the mean and standard deviation over 5 random seeds. We consider two sets of embedding models to construct the multi-view data: \ding{182}~\textbf{OpenAIs}, including 3 OpenAI family models, i.e., OpenAI-small/ada/large, and \ding{183}~\textbf{Mixed}, including 4 representative models, i.e., OpenAI-large, BERT, Qwen, and Llama. The details of implementation and embedding models are given in Appendices~\ref{app:implement} and ~\ref{app:embedding}, respectively.

\subsection{Main Results}
Table~\ref{tab:main_res} shows the comparison results of \ourmethod with baseline methods in terms of AUROC. Results in AUPRC are provided in Appendix~\ref{app:auprc}. We have the following observations. 
\ding{182} \ourmethod achieves the best AUROC on 9/10 datasets and remains competitive on \texttt{SMSSpam}, demonstrating its strong generalization ability across different domains.
\ding{183}~Compared with the strongest baselines, \ourmethod brings clear improvements on challenging datasets. %
These gains indicate that leveraging multi-view signals is more effective than pairing a single strong representation with a fixed detector. 
\ding{184}~\ourmethod consistently outperforms multi-view baselines (NCMOD and RCPMOD) on the same embedding sets. This suggests that our framework is better tailored for multi-view TAD by learning from high-dimensional text embeddings. %
\ding{185} The OpenAIs performs best on most NLP-ADBench datasets, while Mixed is competitive on TAD-Bench and achieves the best result on \texttt{CovidFake} (0.9776). This suggests that mixing heterogeneous backbones can provide complementary cues for domain-specific anomalies.

\begin{table}[t]
\centering
\resizebox{1\columnwidth}{!}{
\begin{tabular}{l c c c c}
\toprule
\textbf{Variants} & \textbf{\shortstack{NLPAD-\\BBCNews}} & \textbf{\shortstack{NLPAD-\\AGNews}} & \textbf{\shortstack{NLPAD-\\MovieReview}} & \textbf{\shortstack{TAD-\\OLID}} \\
\midrule
\multicolumn{5}{c}{\textbf{OpenAIs}} \\
\midrule
\ourmethod & \cellcolor{gray!20}\textbf{0.9860} & \cellcolor{gray!20}\textbf{0.9484} & \cellcolor{gray!20}\textbf{0.8381} & \cellcolor{gray!20}\textbf{0.6355} \\
w/o AA & 0.9858 & \cellcolor{gray!20}\textbf{0.9484} & 0.8378 & 0.6314 \\
w/o CC & 0.9788 & 0.8811 & 0.6592 & 0.5179 \\
w/o AE & 0.9775 & 0.9454 & 0.8350 & 0.6341 \\
\midrule
\multicolumn{5}{c}{\textbf{Mixed}} \\
\midrule
\ourmethod & \cellcolor{gray!20}\textbf{0.9752} & \cellcolor{gray!20}\textbf{0.9482} & \cellcolor{gray!20}\textbf{0.7914} & \cellcolor{gray!20}\textbf{0.4892} \\
w/o AA & \cellcolor{gray!20}\textbf{0.9752} & 0.9480 & 0.7869 & 0.4841 \\
w/o CC & 0.9721 & 0.8917 & 0.5582 & 0.4592 \\
w/o AE & 0.9474 & 0.9432 & 0.7913 & 0.4883 \\
\bottomrule
\end{tabular}}
\caption{Ablation results on AUROC for four datasets.}\label{tab:ablation}

\end{table}

\subsection{Ablation Study}
To validate the contributions of key components in \ourmethod, we construct three variants: \ding{182}~\textbf{w/o AA}, which replaces the adaptive allocation module with uniform fusion over views; \ding{183}~\textbf{w/o CC}, which removes the contrastive collaboration module; and \ding{184}~\textbf{w/o AE}, which removes the autoencoder reconstruction loss and trains the model only with the contrastive learning loss.

The results are summarized in Table~\ref{tab:ablation}, which shows that all components contribute to the final performance consistently. 
\ding{182}~Removing the allocation module results in a small but consistent drop. This indicates that sample-level adaptive view weighting helps when per-sample view quality varies. 
\ding{183}~Removing the contrastive collaboration module causes the largest performance drop on almost all datasets and in both view combinations, e.g., on \texttt{MovieReview} ($-17.89\%$/$-23.32\%$). These results indicate that reconstruction alone overlooks the alignment of heterogeneous views, and the learned scores can become view-specific, which harms anomaly scoring.
\ding{184}~Removing AE yields a moderate decrease, and the effect is more pronounced with heterogeneous views, e.g., on \texttt{BBCNews} in the mixed-backbone setting (0.9752$\rightarrow$0.9474). This shows that AE provides a stabilizer when the views come from more diverse embedding model families, and the contrastive objective alone may overfit to cross-view shortcuts.

\begin{figure}[t]
\centering
\includegraphics[width=1.0\columnwidth]{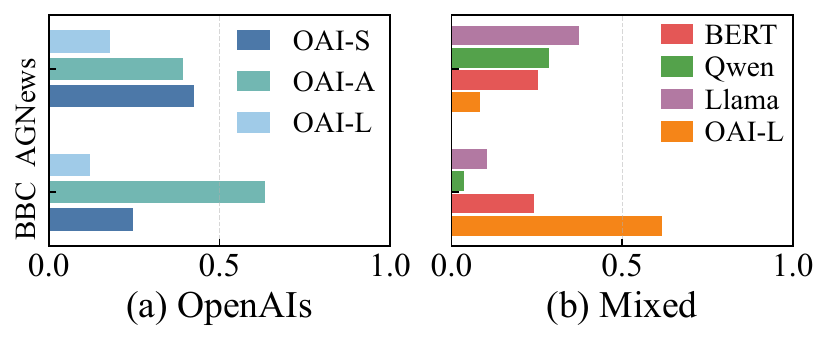}
\caption{Distribution of the top-1 view selected by the gating module on each dataset.}
\label{fig:topview}%
\end{figure}

\subsection{Allocation Visualization}
To gain a deeper understanding of the behavior of the adaptive allocation module, we computed the $\operatorname{argmax}$ of the view weights for each test sample to obtain its top views. We then statistically analyzed the distribution of the top-1 views of all samples in two datasets, with the results shown in Figure~\ref{fig:topview}. The weights exhibit clear dataset-dependent preferences rather than uniformly selecting a fixed view, indicating that the fusion is sample-adaptive. In particular, for the \textbf{OpenAIs} views, \texttt{AGNews} shows a more balanced utilization of small/ada/large, while \texttt{BBCNews} is dominated by OpenAI-ada. For the \textbf{Mixed} views, the dominant view also shifts across datasets (e.g., OpenAI-large is more frequently selected on \texttt{BBCNews}, whereas Llama/BERT/Qwen receive higher selections on \texttt{AGNews}). These results suggest that different views capture complementary cues, and the allocation module can automatically emphasize the most informative views for a given dataset.

\subsection{Robustness Analysis}
\begin{figure}[t]
\centering
\subfloat[BBCNews]{
\label{subfig:bbc_robust}%
\includegraphics[height=0.385\columnwidth]{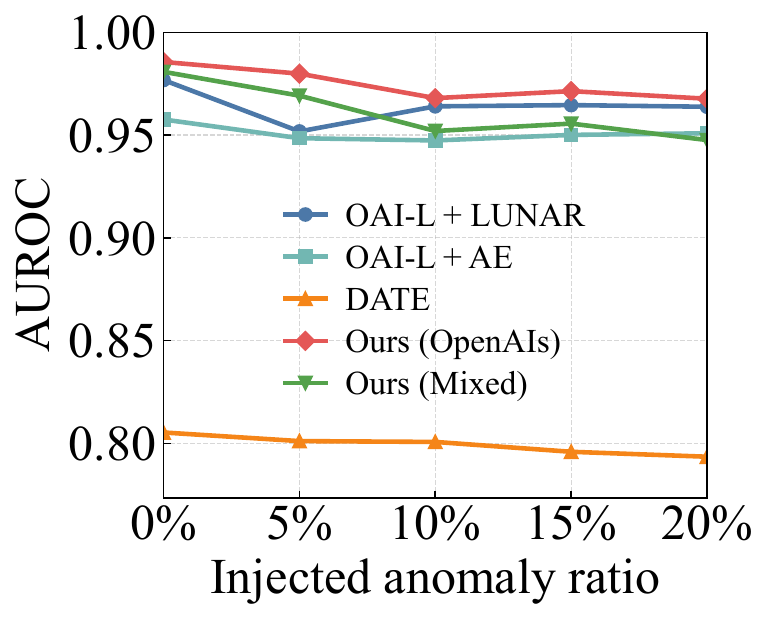}}
\hfill
\subfloat[Liar2]{
\label{subfig:liar2_robust}
\includegraphics[height=0.385\columnwidth]{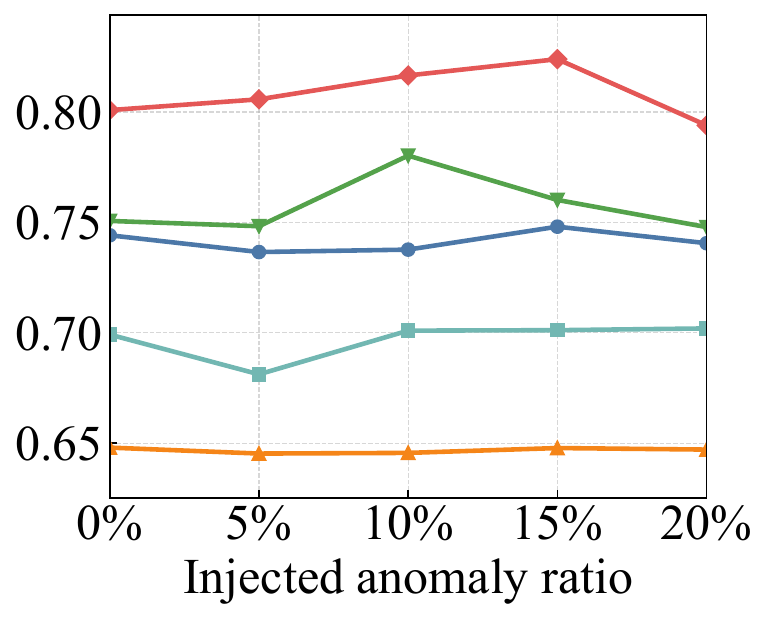}}
\caption{Model robustness under different anomaly inject ratio (\%) in inlier training data.}
\label{fig:robustness}
\end{figure}

To evaluate robustness against contaminated training data, we gradually inject anomalous instances into the inlier-only training set and report AUROC under different injection ratios in Figure~\ref{fig:robustness}. 
Overall, \ourmethod remains stable as the injected anomaly ratio increases, showing only mild performance variations, while maintaining a clear margin over strong baselines such as \texttt{OpenAI-L+LUNAR} and \texttt{OpenAI-L+AE}. 
On \texttt{BBCNews} (Figure~\ref{subfig:bbc_robust}), all methods exhibit relatively small changes, but \ourmethod (OpenAIs) consistently achieves the highest AUROC across all ratios, indicating robustness even when the training set is slightly polluted. 
On \texttt{Liar2} (Figure~\ref{subfig:liar2_robust}), the gap becomes more evident: \ourmethod (OpenAIs) preserves strong performance under increasing contamination, and the Mixed variant remains competitive, suggesting that sample-adaptive multi-view fusion can mitigate the adverse impact of noisy inlier training data.

\subsection{Hyperparameter Analysis}
\begin{figure}[t]
\centering
\subfloat[BBCNews]{
\label{subfig:bbc_hyper}%
\includegraphics[width=0.48\columnwidth]{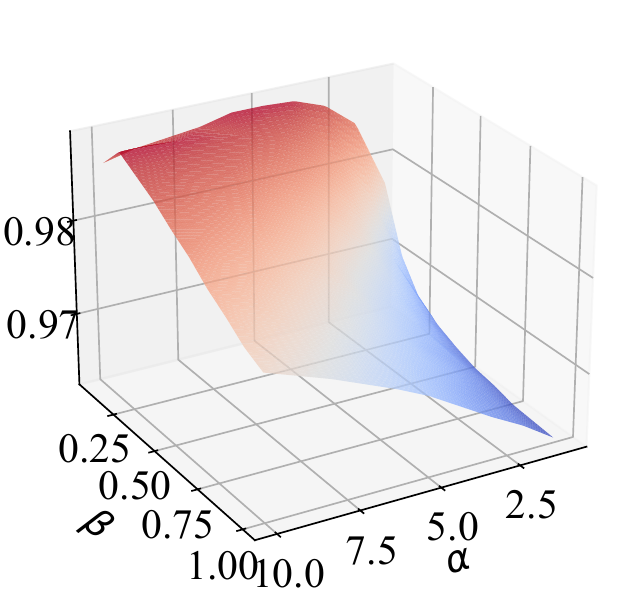}}
\hfill
\subfloat[Liar2]{
\label{subfig:liar_hyper}
\includegraphics[width=0.48\columnwidth]{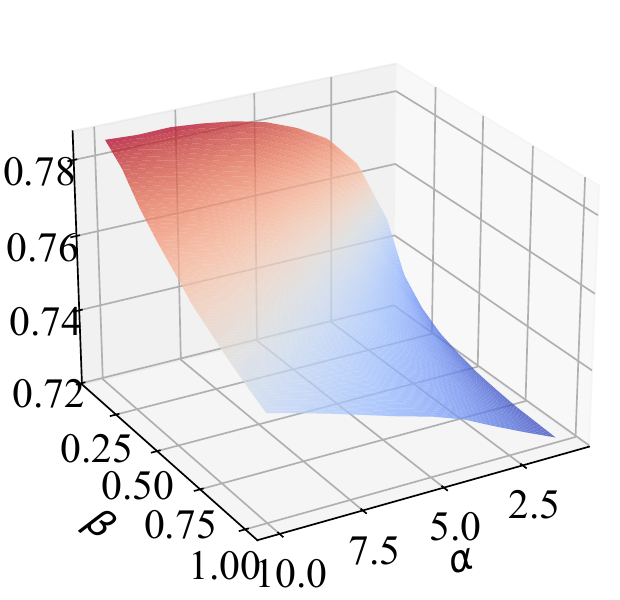}}
\caption{Hyperparameter sensitivity analysis.}
\label{fig:hyperparam}
\end{figure}

We study the sensitivity of \ourmethod to the balance hyperparameters for reconstruction-based and contrastive-based scores, i.e., $\alpha$ and $\beta$. Figure~\ref{fig:hyperparam} illustrates the changes in AUROC as $\alpha$ and $\beta$ vary. Overall, the performance changes smoothly over the grid and preserves a broad high-performing region, indicating that \ourmethod is not overly sensitive to precise hyperparameter tuning. On \texttt{BBCNews} (Figure~\ref{subfig:bbc_hyper}), AUROC remains near-optimal for moderate-to-large $\alpha$, while it drops when $\beta$ becomes too large, suggesting that overweighting the cross-view consistency term can introduce noise into scoring and hurt discrimination. 
On \texttt{Liar2} (Figure~\ref{subfig:liar_hyper}), the trend is more pronounced: the best results are obtained with a sufficiently large $\alpha$ and a small-to-moderate $\beta$, whereas increasing $\beta$ steadily degrades AUROC. In practice, these results suggest prioritizing reconstruction evidence and using the contrastive-based score as a complementary signal.

\section{Conclusion}
In this paper, we propose a novel multi-view TAD method, \ourmethod, that leverages embeddings from multiple language models to advance text anomaly detection (TAD). \ourmethod adopts a multi-view reconstruction model as the backbone, with a contrastive collaboration module to enhance and align the inter-view consistency. Furthermore, we design an adaptive allocation module that automatically assigns appropriate contribution weights to different views for anomaly detection. Extensive experiments demonstrate the state-of-the-art performance of \ourmethod on multiple benchmark datasets and its strong robustness under varying data contamination.


\section*{Limitations}

While \ourmethod demonstrates strong capability in TAD, it currently relies on accessing multiple pretrained embedding models, which may introduce additional inference cost and latency in practical deployments. This dependence on multiple external models may also limit scalability in resource-constrained environments. A promising direction for future work is to explore more efficient strategies that can actively and incrementally request embeddings only from the most informative and suitable models, or dynamically select a subset of views based on task characteristics or runtime constraints. Such adaptive embedding acquisition mechanisms would help further improve the efficiency, scalability, and practicality of multi-view TAD systems.

\section*{Ethical Considerations}

Our research involves no human subjects, animal experiments, or sensitive data. All experiments are conducted using publicly available datasets within simulated environments. We identify no ethical risks or conflicts of interest. We are committed to upholding the highest standards of research integrity and ensuring full compliance with ethical guidelines. Nonetheless, any real-world deployment should safeguard data privacy and carefully manage potential false alarms to prevent bias or discrimination.
\bibliography{ref}

\clearpage

\appendix

\section*{Appendices}

\section{Detailed Related Work} \label{app:rw}
\subsection{Text Embedding}\label{subsec:emb_model}
Text embedding techniques aim to map textual data into vectorized representations (a.k.a. embeddings) that capture semantic and syntactic information. Early methods utilize bag-of-words representations such as TF-IDF~\cite{salton1988term} to encode text as sparse frequency-based vectors. Later, neural embedding methods such as Word2Vec learn dense word representations based on contextual word prediction~\cite{mikolov2013efficient}. With the development of Transformer models, encoder-only models like BERT learn contextualized language representations learned via large-scale pretraining~\cite{devlin2019bert}. 

In the era of LLMs, text embeddings generated by billion-parameter models have become more expressive and powerful in various text-related tasks~\cite{shen-etal-2025-understanding,liu2025graph,li2026assemble,miao2025blindguard,pan2025explainable}. For example, based on GPT architectures, OpenAI provides text embedding models with different scales to meet various application needs~\cite{openai2024embedding}. Likewise, the Qwen series releases multiple text embedding models of different sizes for representation learning~\cite{zhang2025qwen3}. These advanced models provide high-quality embeddings for various downstream tasks, including anomaly detection.

\subsection{Text Anomaly Detection}
Text anomaly detection (TAD) aims to identify textual instances that deviate significantly from dominant normal data~\cite{li2024nlp}. Existing TAD methods can be divided into two categories, i.e., end-to-end methods and embedding-based methods (a.k.a. two-step methods). 

End-to-end methods perform anomaly detection in a unified manner by directly predicting abnormality from raw textual inputs. Early methods employ autoencoder-based reconstruction models to reconstruct normal text patterns and identify anomalies via reconstruction errors~\cite{manevitz2007one}. CVDD utilized distance learning for context vectors to identify samples that deviate from normal patterns as anomalies~\cite{ruff2019self}. Based on Transformer models, DATE utilizes self-supervised learning at both the token level and the sequence level to capture normal textual patterns for TAD~\cite{manolache2021date}. FATE employs a deviation learning technique to build a text anomaly detection model~\cite{das2023few}.

Unlike other modalities where end-to-end anomaly detection methods dominate~\cite{pan2025label,zheng2022unsupervised,liu2024arc,li2026clip}, state-of-the-art performance in text anomaly detection is achieved by embedding-based methods.  Following a two-step pipeline, embedding-based methods first convert text into dense embeddings using pretrained text embedding models and then apply anomaly detectors on the compact embeddings. While the embeddings can be acquired by various models introduced in Section~\ref{subsec:emb_model}, the detection can be conducted by different types of anomaly detection algorithms. Common choices include density‐based methods (e.g., LOF~\cite{breunig2000lof}), distance‐based methods (e.g., kNN~\cite{ramaswamy2000efficient}), statistical methods (e.g., COPOD~\cite{li2020copod}), tree‐based methods (e.g., iForest~\cite{liu2008isolation}), and deep learning–based methods (e.g., DeepSVDD~\cite{ruff2018deep} and LUNAR~\cite{goodge2022lunar}).

Although end-to-end methods are theoretically capable of directly learning anomaly patterns from raw text, empirical evidence in recent benchmarking studies~\cite{li2024nlp,cao2025tad} shows that embedding-based methods often achieve better performance. Nevertheless, existing embedding-based methods typically rely on a single embedding model, which makes them less robust when facing diverse datasets and anomaly types.

\subsection{Multi-View Anomaly Detection}

While conventional anomaly detection usually operates on a single view of the data~\cite{pan2025survey,zhao2025freegad,pan2026correcting}, multi-view anomaly detection focuses on identifying anomalous samples in multi-view data, e.g., image data represented by multiple views like color and shape feature descriptors. Early studies detect anomalies by clustering data and identifying samples that deviate from the learned clusters~\cite{marcos2013clustering,liu2012using}. Taking advantage of deep learning, NCMOD applies an autoencoder to learn a latent representation of the data and constructs neighborhood consensus graphs~\cite{li2024noise} to detect outliers~\cite{cheng2021neighborhood}. \cite{tian2024multi} propose to aligns instance–neighborhood structures and perform cross-view reasoning to better detect inconsistent anomalies across views. RCPMOD utilizes contrastive regularization~\cite{tan2025bisecle} and neighbor-based completion to detect anomalies in partial multi-view data~\cite{wang2024regularized}. Despite their success in multi-view visual data, how to conduct multi-view anomaly detection for high-dimensional textual data remains an open problem. 

\section{Algorithm Description} \label{app:algo}
The training and testing algorithms are given in Algorithm~\ref{algo:train} and Algorithm~\ref{algo:test}, respectively.
\begin{algorithm}[ht]
\caption{The Training Algorithm of \ourmethod}
\label{algo:train}
\LinesNumbered
\KwIn{Normal-only training set $\mathcal{D}_{\text{train}}=\{x_i\}_{i=1}^N$; number of views $K$.}
\Param{$E$; $\lambda,\tau$.}
Compute multi-view embeddings $\mathbf{v}_i^{(k)}=f_k(x_i)$ via Eq.~\eqref{eq:mv_embed}.\\
Fit view aligners on $\mathbf{V}^{(k)}$ with PCA.\\
Initialize model parameters.\\
\textbf{Stage 1: Detection model training.}\\
Freeze the allocation module and enforce uniform weights $w_i^{(k)}\!=\!1/K$.\\
\For{$e = 1:E$}{
For $k=1,\dots,K$ and $i=1,\dots,N$, obtain  $\mathbf{z}_i^{(k)}$ and $\widehat{\mathbf{v}}_i^{(k)}$ via Eq.~\eqref{eq:ae}.\\
Compute $\mathcal{L}_{i,rec}^{(k)}$ via Eq.~\eqref{eq:l_rec} and $\mathcal{L}_{i,con}^{(j,k)}$ via Eq.~\eqref{eq:l_con}.\\
Update the backbone parameters by minimizing the overall loss via Eq.~\eqref{eq:loss}.\\
}
\textbf{Stage 2: Allocation learning.}\\
Freeze the backbone parameters.\\
\For{$e = 1:E$}{
Compute view-wise anomaly scores $s_{i,\text{rec}}^{(k)}$ and $s_{i,\text{con}}^{(k)}$ via Eq.~\eqref{eq:s_rec} and Eq.~\eqref{eq:s_con}.\\
Obtain aligned features $\widetilde{\mathbf{v}}_i^{(k)}$ by applying the fitted PCA on $\mathbf{v}_i^{(k)}$.\\
Compute allocation weights $w_i^{(k)}$ via Eq.~\eqref{eq:weight}.\\
Update the allocation module by minimizing  Eq.~\eqref{eq:loss}.\\
}
\end{algorithm}

\begin{algorithm}[ht]
\caption{The Inference algorithm of \ourmethod}
\label{algo:test}
\LinesNumbered
\KwIn{Test dataset $\mathcal{D}_{\text{test}}$; number of views $K$.}
\Param{Well-trained model weight parameters; $\alpha,\beta,\tau$.}

\ForEach{$x\in\mathcal{D}_{\text{test}}$}{
Compute multi-view embeddings $\mathbf{v}^{(k)}=f_k(x)$ via Eq.~\eqref{eq:mv_embed}.\\
Obtain $\mathbf{z}^{(k)}$ and $\widehat{\mathbf{v}}^{(k)}$ via Eq.~\eqref{eq:ae}.\\
Compute view-wise anomaly scores $s_{\text{rec}}^{(k)}$ and $s_{\text{con}}^{(k)}$ via Eq.~\eqref{eq:s_rec} and Eq.~\eqref{eq:s_con}.\\
Obtain aligned features $\widetilde{\mathbf{v}}^{(k)}$ by applying the fitted PCA on $\mathbf{v}^{(k)}$.\\
Compute allocation weights $w^{(k)}$ via Eq.~\eqref{eq:weight}.\\
Return the final anomaly score $s(x)$ via Eq.~\eqref{eq:final_score}.\\
}
\end{algorithm}

\section{Complexity Analysis} \label{app:complex}
In this subsection, we discuss the time complexity of \ourmethod in the testing phase. The overall cost mainly consists of embedding extraction, view alignment, backbone inference, and scoring.
The complexity of embedding extraction is $\mathcal{O}\big(N\sum_{k=1}^{K} C_{\text{emb}}^{(k)}\big)$, where $N$ is the number of samples, $K$ is the number of views, and $C_{\text{emb}}^{(k)}$ denotes the per-sample cost of extracting view-$k$ embeddings (i.e., one call to $f_k$).
The complexity of feature alignment (PCA projection) is $\mathcal{O}\big(N\sum_{k=1}^{K} d_k d\big)$, where $d_k$ is the embedding dimension of view $k$ and $d$ is the aligned dimension.
The complexity of backbone inference is $\mathcal{O}\big(N\sum_{k=1}^{K} C_{\text{ae}}^{(k)}\big)$, where $C_{\text{ae}}^{(k)}$ denotes the per-sample forward cost of the view-$k$ autoencoder in Eq.~\eqref{eq:ae}.
The complexity of weight estimation in Eq.~\eqref{eq:weight} is $\mathcal{O}(N K d)$, where $d$ is the aligned feature dimension fed into the MLP-based estimator.
The complexity of score fusion in Eq.~\eqref{eq:final_score} is $\mathcal{O}(N K)$.
To sum up, the overall testing complexity is $\mathcal{O}\big(N\sum_{k=1}^{K} C_{\text{emb}}^{(k)} + N\sum_{k=1}^{K} d_k d + N\sum_{k=1}^{K} C_{\text{ae}}^{(k)} + N K d + N K\big)$, which is approximately linear in $N$ when $K$, $d$, and $d_k$ are treated as constants.

\section{Datasets} \label{app:dset}
We conduct experiments on 10 text anomaly detection datasets spanning multiple domains: four news-related datasets (NLPAD-AGNews, NLPAD-BBCNews, NLPAD-N24News, and TAD-COVIDFake), two spam filtering datasets (TAD-EmailSpam and TAD-SMSSpam), one review sentiment dataset (NLPAD-MovieReview), and three social media content datasets (TAD-Liar2, TAD-OLID, and TAD-HateSpeech). Following the standard anomaly detection protocol, we designate a specific class as anomalous and apply downsampling to create class imbalance. Dataset statistics are summarized in Table~\ref{app:dataset}, with detailed descriptions provided below:

\begin{itemize}
    \item \textbf{NLPAD-AGNews} derives from the AG News corpus~\cite{rai2023agnews}, a benchmark originally designed for classifying news articles into topics. This corpus encompasses 127,600 samples spanning four categories: World, Sports, Business, and Sci/Tech. We extract textual content from the ``description'' field to construct our dataset, treating the ``World'' category as anomalous instances with appropriate downsampling.
    \item \textbf{NLPAD-BBCNews} builds upon the BBC News corpus~\cite{han2022adbench}, initially developed for multi-topic document classification. The corpus comprises 2,225 news articles spanning five categories: Business, Entertainment, Politics, Sport, and Tech. We utilize the complete article text as input, with the ``Entertainment'' category serving as the anomalous class after downsampling.
    \item \textbf{NLPAD-MovieReview} originates from the Movie Review corpus~\cite{manolache2021date}, a widely-adopted benchmark for sentiment classification of film reviews. This corpus contains 50,000 reviews with binary sentiment labels representing positive and negative sentiments. We employ the complete review text, designating the ``negative'' category as the anomalous class with corresponding downsampling.
    \item \textbf{NLPAD-N24News} is derived from the N24News corpus~\cite{wang2022n24news}, originally curated for categorizing news content by topic. The corpus encompasses 61,235 articles distributed across multiple categories. We leverage the full article text, with the ``food'' category treated as anomalous after appropriate downsampling.
    \item \textbf{TAD-EmailSpam} originates from the Spam Emails corpus~\cite{metsis2006spam}, a benchmark widely adopted for identifying unsolicited email messages. This corpus comprises 5,171 email samples with binary labels distinguishing spam from legitimate correspondence. We extract content from email body text to construct our dataset, designating the ``spam'' category as the anomalous class with corresponding downsampling.
    \item \textbf{TAD-SMSSpam} derives from the SMS Spam Collection corpus~\cite{almeida2011contributions}, initially developed for filtering unwanted text messages. The corpus encompasses 5,574 SMS messages with binary classifications distinguishing spam from legitimate messages. We utilize the complete message text as input, treating the ``spam'' category as anomalous after appropriate downsampling.

    \begin{table}[t]
\centering
\resizebox{1\columnwidth}{!}{
\begin{tabular}{l | c c c c}
\toprule
\textbf{Dataset} & \textbf{\shortstack{\#Samples}} & \textbf{\shortstack{\#Normal}} & \textbf{\shortstack{\#Anomaly}} & \textbf{\shortstack{\%Anomaly}} \\
\midrule
 NLPAD-AGNews & 98,207 & 94,427 & 3,780 & 3.85\% \\
 NLPAD-BBCNews & 1,785 & 1,723 & 62 & 3.47\% \\
 NLPAD-MovieReview & 26,369 & 24,882 & 1,487 & 5.64\% \\
 NLPAD-N24News & 59,822 & 57,994 & 1,828 & 3.06\% \\
 TAD-EmailSpam & 3,578 & 3,432 & 146 & 4.08\% \\
 TAD-SMSSpam & 4,969 & 4,825 & 144 & 2.90\% \\
 TAD-OLID & 641 & 620 & 21 & 3.28\% \\
 TAD-HateSpeech & 4,287 & 4,163 & 124 & 2.89\% \\
 TAD-CovidFake & 1,173 & 1,120 & 53 & 4.52\% \\
 TAD-Liar2 & 2,130 & 2,068 & 62 & 2.91\% \\
\bottomrule
\end{tabular}}
\caption{Statistical of datasets.}
\label{app:dataset}
\end{table}

    \begin{table*}[t]
\centering
\resizebox{\textwidth}{!}{
\begin{tabular}{lcccccccc}
\toprule
\textbf{Dataset} & \textbf{\shortstack{Stage-1\\Epochs}} & \textbf{\shortstack{Stage-2\\Epochs}} & \textbf{\shortstack{Backbone LR}} & \textbf{\shortstack{Allocation LR}} & \textbf{\shortstack{Batch Size}} & \textbf{$\lambda$} & \textbf{$\alpha$} & \textbf{$\beta$} \\
\midrule
NLPAD-AGNews & 100 & 15 & 1e-3 & 1e-4 & 256 & 2e-2 & 1 & 0.1 \\
NLPAD-BBCNews & 100 & 50 & 1e-3 & 1e-3 & Full-Batch & 2e-2 & 1 & 0.1 \\
NLPAD-MovieReview & 30 & 50 & 1e-3 & 1e-3 & 256 & 1 & 1 & 1 \\
NLPAD-N24News & 15 & 5 & 1e-3 & 1e-3 & 256 & 1 & 1 & 5 \\
TAD-EmailSpam & 45 & 50 & 1e-3 & 1e-3 & 256 & 2e-2 & 5 & 0.1 \\
TAD-SMSSpam & 30 & 50 & 1e-3 & 1e-3 & 256 & 1 & 1 & 1 \\
TAD-OLID & 20 & 50 & 1e-2 & 1e-3 & 256 & 20 & 1 & 1 \\
TAD-HateSpeech & 10 & 50 & 1e-2 & 1e-3 & 256 & 1 & 1 & 1 \\
TAD-CovidFake & 80 & 50 & 2e-3 & 1e-3 & Full-Batch & 2e-2 & 1 & 0.1 \\
TAD-Liar2 & 100 & 50 & 1e-3 & 1e-2 & Full-Batch & 2e-2 & 5 & 0.1 \\
\bottomrule
\end{tabular}}
\caption{Searched hyper-parameters for each benchmark dataset.}
\label{tab:app_hyper}
\end{table*}
    \item \textbf{TAD-CovidFake} derives from the COVID-Fake corpus~\cite{das2021heuristic}, originally developed for distinguishing authentic COVID-19 information from misinformation. This corpus comprises 10,700 samples aggregating social media posts and fact-checked content from diverse sources with binary labels distinguishing fake from real news content. We utilize the complete textual content as input, treating the ``fake'' category as anomalous after appropriate downsampling.

    \item \textbf{TAD-Liar2} originates from a fact-checking corpus~\cite{xu2023comparative}, a benchmark initially designed for veracity assessment of public claims. The corpus encompasses approximately 23,000 statements annotated by expert fact-checkers across multiple veracity levels. We extract the claim text to construct our dataset, designating the ``Pants on Fire'' category as the anomalous class with corresponding downsampling.
    \item \textbf{TAD-OLID} builds upon the Offensive Language Identification Dataset corpus~\cite{zampieri2019predicting}, originally curated for detecting offensive content in social media. This corpus comprises 14,200 English tweets with hierarchical annotations spanning three classification levels. We leverage Level A annotations utilizing the complete tweet text, with the ``offensive'' category treated as anomalous after appropriate downsampling.
    \item \textbf{TAD-HateSpeech} is derived from a crowdsourced corpus~\cite{davidson2017automated}, initially developed for identifying hate speech in Twitter content. The corpus encompasses 25,296 tweets annotated through CrowdFlower with classifications distinguishing hate speech from offensive language and neutral content. We employ the complete tweet text as input, treating the ``hate speech'' category as the anomalous class with corresponding downsampling.
\end{itemize}

\begin{table}[t]
\centering
\begin{tabular}{l c}
\toprule
\textbf{Parameter} & \textbf{Value} \\
\midrule
Latent dimension & 128 \\
Encoder hidden dimensions & [512, 256] \\
Decoder hidden dimensions & [256, 512] \\
Batch normalization & True \\
Activation function & ReLU \\
Decoder final activation & Sigmoid \\
PCA projection dimension & 128 \\
Allocation activation function & Sigmoid \\
Optimizer & Adam \\
Weight decay & 0 \\
Contrastive temperature & 0.5 \\
\bottomrule
\end{tabular}
\caption{Fixed hyper-parameters.}
\label{tab:app_fixed_hyper}
\end{table}
\section{Implementation Details} \label{app:implement}
\textbf{Hyper-parameters.} 
We employ a systematic hyperparameter tuning approach to optimize model performance, focusing on parameters that significantly influence outcomes while maintaining fixed values for those with minimal impact. The optimal hyperparameter configurations for each benchmark dataset are presented in Table~\ref{tab:app_hyper}, whereas the fixed parameter settings are detailed in Table~\ref{tab:app_fixed_hyper}. Our hyperparameter optimization systematically explores the following search space:

\begin{itemize}
\item Stage-1 training epochs: \{10, 15, 30, 45, 50, 80, 100, 200\}
\item Stage-2 training epochs: \{5, 15, 30, 50\}
\item Backbone learning rate: \{1e-4, 1e-3, 2e-3, 1e-2\}
\item Allocation learning rate: \{1e-5, 1e-4, 1e-3, 1e-2\}
\item Batch size: \{256, 512, 1024, Full-Batch\}
\item Loss weight $\lambda$: \{0.01, 0.02, 0.1, 1, 10, 20\}
\item Reconstruction score weight $\alpha$: \{1, 2, 3, 4, 5, 6, 7, 8, 9, 10\}
\item Contrastive score weight $\beta$: \{0.1, 0.2, 0.3, 0.4, 0.5, 0.6, 0.7, 0.8, 0.9, 1.0\}
\end{itemize}

To ensure robust and reliable results, we also conducted a comprehensive grid search to obtain the best hyperparameter configurations for the baselines. Specifically, for RCPMOD~\cite{wang2024regularized}, we performed grid searches on model-specific hyperparameters including the k-nearest neighbor loss weight, memory bank size for contrastive learning, and the number of neighbors for local structure modeling. Similarly, for NCMOD~\cite{cheng2021neighborhood}, we conducted grid searches on the k-nearest neighbor parameters that control the local neighborhood relationships for anomaly score computation.

\noindent\textbf{Computing infrastructures.} We implement the proposed method with Python 3.9 and PyTorch 2.8.0. The key dependencies include scikit-learn, Pandas, sentence-transformers, transformers, and Numpy. All experiments are conducted on an Ubuntu server with Intel Xeon Platinum 8352V CPU (16 vCPU) and NVIDIA RTX 4090 GPU (24GB) with CUDA 11.8.

\begin{table}[t]
\centering
\resizebox{1\columnwidth}{!}{
\begin{tabular}{l c c c}
\toprule
\textbf{Models} & \textbf{Max Tokens} & \textbf{\# Dimensions} & \textbf{\# Parameters} \\
\midrule
BERT & 512 & 768 & 110 M \\
OAI-A & 8,191 & 1,536 & N/A \\
OAI-S & 8,191 & 1,536 & N/A \\
OAI-L & 8,191 & 3,072 & N/A \\
LLAMA & 4,096 & 2,048 & 1.24 B \\
Qwen & 8,192 & 1,536 & 1.54 B \\
\bottomrule
\end{tabular}}
\caption{Embedding Models Overview. M and B are for million and billion, respectively.}
\label{tab:app_embedding}
\end{table}

\begin{table*}[t]
\centering
\resizebox{1\linewidth}{!}{%
\begin{tabular}{l c c c c c c c c c c}
\toprule
\textbf{Methods} & \textbf{\shortstack{NLPAD-\\AGNews}} & \textbf{\shortstack{NLPAD-\\BBCNews}} & \textbf{\shortstack{NLPAD-\\MovieReview}} & \textbf{\shortstack{NLPAD-\\N24News}} & \textbf{\shortstack{TAD-\\EmailSpam}} & \textbf{\shortstack{TAD-\\SMSSpam}} & \textbf{\shortstack{TAD-\\OLID}} & \textbf{\shortstack{TAD-\\HateSpeech}} & \textbf{\shortstack{TAD-\\CovidFake}} & \textbf{\shortstack{TAD-\\Liar2}} \\
\midrule
CVDD & 0.1652 & 0.1633 & 0.1595 & 0.1847 & 0.4286 & 0.0756 & 0.1125 & 0.0908 & 0.6573 & 0.1608 \\
DATE & 0.3332 & 0.4005 & 0.1575 & 0.2091 & 0.8438 & 0.7038 & 0.1108 & 0.1383 & 0.4049 & 0.2386 \\
FATE & 0.7367 & 0.3164 & 0.2655 & 0.7231 & 0.2804 & \cellcolor{gray!20}\textbf{0.7414} & 0.0784 & 0.1368 & 0.6239 & 0.1288 \\
\midrule
BERT+LOF & 0.2547 & 0.5974 & 0.1617 & 0.1673 & 0.2474 & 0.1304 & 0.0986 & 0.0827 & 0.6853 & 0.1678 \\
BERT+DeepSVDD & 0.1342 & 0.1703 & 0.1448 & 0.0821 & 0.1875 & 0.1140 & 0.1088 & 0.0952 & 0.3897 & 0.1434 \\
BERT+ECOD & 0.1615 & 0.1981 & 0.1372 & 0.0927 & 0.2059 & 0.0982 & 0.0970 & 0.0887 & 0.5703 & 0.1265 \\
BERT+iForest & 0.1630 & 0.2000 & 0.1366 & 0.0898 & 0.1904 & 0.1015 & 0.0961 & 0.0903 & 0.5496 & 0.1247 \\
BERT+SO-GAAL & 0.1055 & 0.0786 & 0.1500 & 0.0917 & 0.1117 & 0.0705 & 0.1067 & 0.0995 & 0.4006 & 0.1212 \\
BERT+AE & 0.2211 & 0.4214 & 0.1477 & 0.1254 & 0.3046 & 0.1517 & 0.0977 & 0.0859 & 0.6775 & 0.1501 \\
BERT+VAE & 0.1884 & 0.2469 & 0.1401 & 0.0981 & 0.2263 & 0.1135 & 0.0981 & 0.0893 & 0.5837 & 0.1357 \\
BERT+LUNAR & 0.2655 & 0.6097 & 0.1484 & 0.1433 & 0.3730 & 0.1539 & 0.1056 & 0.0924 & 0.6862 & 0.1472 \\
OAI-L+LOF & 0.2923 & 0.7693 & 0.3129 & 0.2059 & 0.4383 & 0.1810 & 0.1538 & 0.1066 & 0.5700 & 0.2391 \\
OAI-L+DeepSVDD & 0.1219 & 0.1356 & 0.1690 & 0.1368 & 0.1853 & 0.0821 & 0.1391 & 0.0971 & 0.1769 & 0.0907 \\
OAI-L+ECOD & 0.1918 & 0.2328 & 0.1532 & 0.1303 & 0.5544 & 0.0709 & 0.0999 & 0.0631 & 0.5862 & 0.1352 \\
OAI-L+iForest & 0.1496 & 0.1564 & 0.1700 & 0.1158 & 0.4914 & 0.0878 & 0.1085 & 0.0830 & 0.4126 & 0.1172 \\
OAI-L+SO-GAAL & 0.0871 & 0.0653 & 0.1391 & 0.0618 & 0.0724 & 0.0791 & 0.1048 & 0.0630 & 0.0916 & 0.0998 \\
OAI-L+AE & 0.5132 & 0.7613 & 0.2007 & 0.3013 & 0.6072 & 0.0816 & 0.1071 & 0.1052 & 0.7843 & 0.2182 \\
OAI-L+VAE & 0.3873 & 0.2488 & 0.1654 & 0.1947 & 0.5557 & 0.0726 & 0.0993 & 0.0630 & 0.6144 & 0.1392 \\
OAI-L+LUNAR & 0.6206 & 0.8722 & 0.4303 & 0.4291 & \cellcolor{gray!20}\textbf{0.8815} & 0.1419 & 0.1138 & 0.1500 & 0.8138 & 0.2543 \\
\midrule
NCMOD (\textbf{OpenAIs}) & 0.2411 & 0.3548 & 0.2569 & 0.1929 & 0.6886 & 0.3323 & \cellcolor{gray!20}\textbf{0.1581} & 0.0838 & 0.4848 & 0.2160 \\
NCMOD (\textbf{Mixed}) & 0.2469 & 0.2490 & 0.1971 & 0.1523 & 0.5070 & 0.1101 & 0.1060 & 0.0928 & 0.6741 & 0.1872 \\
\midrule
RCPMOD (\textbf{OpenAIs}) & 0.2935 & 0.8669 & 0.5891 & 0.9313 & 0.4260 & 0.2575 & 0.1408 & 0.1560 & 0.5799 & 0.1992 \\
RCPMOD (\textbf{Mixed}) & 0.4168 & 0.8434 & 0.4052 & 0.9082 & 0.3591 & 0.2161 & 0.1087 & 0.1031 & 0.7012 & 0.1929 \\
\midrule
\ourmethod (\textbf{OpenAIs}) & \cellcolor{gray!20}\textbf{0.9352} & \cellcolor{gray!20}\textbf{0.9160} & \cellcolor{gray!20}\textbf{0.6356} & \cellcolor{gray!20}\textbf{0.9591} & 0.8772 & 0.4433 & 0.1571 & \cellcolor{gray!20}\textbf{0.1805} & 0.7326 & \cellcolor{gray!20}\textbf{0.2937} \\
\ourmethod (\textbf{Mixed}) & 0.9242 & 0.8362 & 0.4869 & 0.9454 & 0.8107 & 0.2568 & 0.1218 & 0.1101 & \cellcolor{gray!20}\textbf{0.9052} & 0.2196 \\
\bottomrule
\end{tabular}}
\caption{Main results on AUPRC. Best results are highlighted in bold and shaded.}
\label{tab:main_res_auprc}
\end{table*}

\section{Text Embedding Models} \label{app:embedding}

To capture the semantic characteristics of textual data in our anomaly detection framework, we employ a collection of state-of-the-art pre-trained embedding models that convert raw text into high-dimensional vector representations. These embedding models form the foundation of our multiview approach, where each model provides a distinct perspective on the textual content. Table~\ref{tab:app_embedding} presents a comprehensive overview of the embedding architectures utilized in our experimental evaluation.

\textbf{BERT\footnote{https://huggingface.co/google-bert/bert-base-uncased} \textit{(bert-base-uncased)}:} BERT is a transformer-based language model that revolutionized natural language understanding through its bidirectional training approach. Unlike traditional language models that process text sequentially, BERT simultaneously considers both left and right context, enabling richer semantic representations. The base-uncased variant processes lowercase text and contains 12 transformer layers with 768 hidden dimensions.

\textbf{OAI-A\footnote{https://platform.openai.com/docs/models/text-embedding-ada-002} \textit{(text-embedding-ada-002)}:} The OAI-A model represents OpenAI's second-generation embedding architecture, designed for high-quality semantic similarity tasks. This model demonstrates strong performance across diverse text understanding benchmarks and supports significantly longer input sequences compared to earlier transformer models. It produces 1536-dimensional embeddings optimized for retrieval and similarity matching applications.

\textbf{OAI-S\footnote{https://platform.openai.com/docs/models/text-embedding-3-small} \textit{(text-embedding-3-small)}:} The OAI-S model is part of OpenAI's third-generation embedding family, offering improved efficiency while maintaining competitive performance. This model provides an optimal balance between computational cost and representation quality, making it suitable for large-scale text analysis tasks where resource constraints are a consideration.

\textbf{OAI-L\footnote{https://platform.openai.com/docs/models/text-embedding-3-large} \textit{(text-embedding-3-large)}:} The OAI-L model represents the flagship embedding model in OpenAI's third-generation series, delivering superior semantic understanding through its expanded parameter space and enhanced training methodology. With 3072-dimensional output vectors, this model captures fine-grained semantic distinctions and excels in complex text understanding scenarios.

\textbf{LLAMA\footnote{https://huggingface.co/meta-llama/Llama-3.2-1B} \textit{(Llama-3.2-1B)}:} The Llama-3.2-1B model is a compact yet powerful language model from Meta's Llama family, specifically optimized for efficient inference while maintaining strong language understanding capabilities. Despite its relatively smaller parameter count, this model demonstrates robust performance in text representation tasks and supports moderate-length input sequences with 2048-dimensional embeddings.

\textbf{Qwen\footnote{https://huggingface.co/Qwen/Qwen2.5-1.5B} \textit{(Qwen2.5-1.5B)}:} Qwen is a multilingual large language model developed by Alibaba Cloud, featuring enhanced performance in both English and Chinese text understanding. This model incorporates advanced training techniques and architectural improvements that enable effective semantic representation across diverse linguistic contexts, producing 1536-dimensional embeddings with extended context length support.

\section{More Experimental Results} \label{app:auprc}

Table~\ref{tab:main_res_auprc} shows the comparison results in terms of AUPRC. We have the following observations. \ding{182} \ourmethod achieves the best AUPRC on 7/10 datasets, demonstrating consistent superiority across different evaluation metrics. On the remaining datasets, \ourmethod still maintains competitive performance compared to specialized baselines. \ding{183} Compared with the strongest baseline on each dataset, \ourmethod shows substantial improvements on several challenging datasets, e.g., \texttt{AGNews} (0.9352 vs 0.7367, +19.85\%), \texttt{BBCNews} (0.9160 vs 0.8722, +4.38\%), and \texttt{N24News} (0.9591 vs 0.9313, +2.78\%). These significant gains highlight the effectiveness of our multi-view approach in precision-recall trade-offs. Moreover, \ourmethod consistently outperforms multi-view baselines (NCMOD and RCPMOD) across most datasets, e.g., on \texttt{AGNews} (0.9352 vs 0.2411/0.2935) and \texttt{MovieReview} (0.6356 vs 0.2569/0.5891). \ding{184} The OpenAIs view set demonstrates superior performance on most datasets, particularly excelling on news-related tasks such as \texttt{BBCNews} (0.9160) and \texttt{N24News} (0.9591). However, the Mixed view set achieves the best result on \texttt{CovidFake} (0.9052), reinforcing that heterogeneous embedding combinations can capture domain-specific anomaly patterns more effectively for certain tasks.

\end{document}